\documentclass[11pt,a4paper]{article}
\usepackage[hyperref]{eacl2021}
\usepackage{times}
\usepackage{latexsym}

\usepackage{microtype}

\usepackage{url}
\usepackage{amsfonts, amssymb}       
\usepackage{nicefrac}       
\usepackage{microtype}      
\usepackage{graphicx}
\usepackage{capt-of}
\usepackage{float}
\usepackage{macros}
\usepackage{mathtools}
\usepackage{array}
\usepackage{url}
\usepackage{multirow}
\usepackage{wrapfig}
\usepackage{amsmath}

\aclfinalcopy 

\title{On-Device Text Representations Robust To Misspellings via Projections}

\author{Chinnadhurai Sankar \\
  Mila, Universit\'e de Montr\'eal\Thanks{Work done during internship at Google}\\
  Montreal, QC, Canada\\
  \texttt{chinnadhurai@gmail.com} \\\And
 
  Sujith Ravi  \\
  Amazon Alexa\Thanks{Work done while at Google AI} \\
  Sunnyvale, CA, USA  \\
  \texttt{sravi@sravi.org}\\ \And
  
  Zornitsa Kozareva  \\
  Google \\
   Mountain View, CA, USA \\
  \texttt{zornitsa@kozareva.com} \\}

\date{}

\begin{document}
\maketitle
\begin{abstract}
Recently, there has been a strong interest in developing natural language applications that live on personal devices such as mobile phones, watches and IoT with the objective to preserve user privacy and have low memory. Advances in Locality-Sensitive Hashing (LSH)-based projection networks have demonstrated state-of-the-art performance in various classification tasks without explicit word (or word-piece) embedding lookup tables by computing on-the-fly text representations. 

In this paper, we show that the projection based neural classifiers are inherently robust to misspellings and perturbations of the input text. We empirically demonstrate that the LSH projection based classifiers are more robust to common misspellings compared to BiLSTMs (with both word-piece \& word-only tokenization) and fine-tuned BERT based methods. When subject to misspelling attacks, LSH projection based classifiers had a small average accuracy drop of 2.94\% across multiple classifications tasks, while the fine-tuned BERT model accuracy had a significant drop of 11.44\%.
\end{abstract}

\section{Introduction}
At the core of Natural Language Processing (NLP) neural models are pre-trained word embeddings like Word2Vec \cite{MikolovT2013}, GloVe \cite{glove} and ELMo \cite{elmo}. They help initialize the neural models, lead to faster convergence and have improved performance for numerous application such as Question Answering \cite{P18-1157}, Summarization \cite{P16-1046}, Sentiment Analysis \cite{D17-1056}. While word embeddings are powerful in unlimited constraints such as computation power and compute resources, it becomes challenging to deploy them to on-device due to their huge size. 

\begin{figure}[!htbp]
\centering
\includegraphics[scale=0.35]{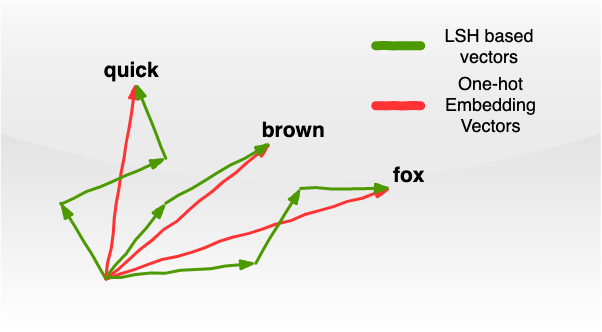}
\caption{One-hot word embedding look-up vectors vs linear combination of LSH projection based vectors \citep{projection_net2017} representing the same word.}
\label{fig:word_embedding_table_vector}
\end{figure}

This led to interesting research by \cite{sgnn_emnlp18,sankar-etal-2019-transferable}, who showed that word embeddings can be replaced with lightweight binary Locality-Sensitive Hashing (LSH) based projections learned on-the-fly.
The projection approach surmounts the need to store any embedding matrices, since the projections are dynamically computed. This further enables user {\it privacy} by performing inference directly on device without sending user data (e.g., personal information) to the server. 
The embedding memory size is reduced from $O(V)$ to $O(K)$, where $V$ is the token vocabulary size and $K << V$, is the binary LSH projection size. The projection representations can operate on either word or character level, and can be used to represent a sentence or a word depending on the NLP application. For instance, recently the Projection Sequence Networks (ProSeqo) \citep{proseqo2019} used BiLSTMs over word-level projection representations to represent long sentences and achieved close to state-of-the-art results in both short and long text classification tasks with varying amounts of supervision and vocabulary sizes.

\begin{figure}[!htbp]
\centering
\includegraphics[scale=0.40]{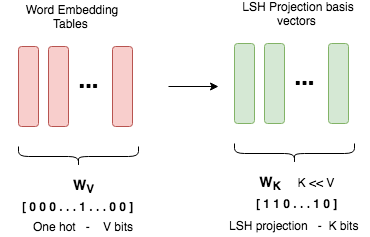}
\caption{Memory for $V$ look-up vectors for each token vs storing $K (<< V)$ vectors and linearly combining them for token representation. We consider $K=1120$ following \citep{sgnn_emnlp18} in this paper.}
\label{fig:word_embedding_table_vs_lsh}
\end{figure}

Despite being successful, there are no existing systematic research efforts focusing on evaluating the capabilities of the LSH based projection for text representations. To that end, we empirically analyze the effectiveness and robustness of the LSH projection approach for text representation by conducting two types of studies in this paper.

\begin{enumerate}
\item \textit{Classification with perturbed inputs}, where we show that Projection based networks 1) Projection Sequence Networks (ProSeqo) \citep{proseqo2019} and 2) Self-Governing Neural Networks (SGNN) models \cite{ravi-kozareva-2019-device} evaluated with perturbed LSH projections are robust to misspellings and transformation attacks, while we observe significant drop in performance for BiLSTMs and fine-tuned BERT classifiers.

\item \textit{Perturbation Analysis}, where we test the robustness of the projection approach by directly analyzing the changes in representations when the input words are subject to the char misspellings.
The purpose of this study is to examine if the words or sentences with misspelling are nearby in the projection space instead of frequently colliding with the projection representations of other valid words.
\end{enumerate}

Overall, our studies showcase the robustness of LSH projection representations and resistance to misspellings. Due to their effectiveness, we believe that in the future, text representations using LSH projections can go beyond memory constrained settings and even be exploited in large scale models like Transformers \citep{vaswani2017attention}.

\begin{figure}[!htbp]
\centering
\includegraphics[scale=0.32]{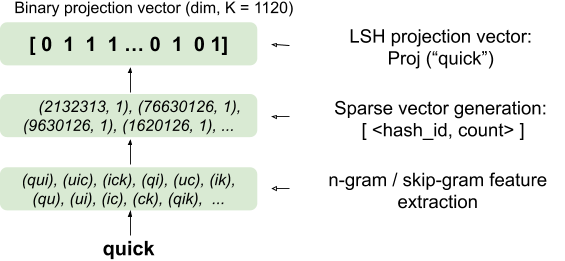}
\caption{Binary Locality-Sensitive Hashing (LSH) projection representation for text.}
\label{fig:lsh_proj}
\end{figure}

\section{Binary LSH projections for text representations} \label{sec:background}
The dependency on vocabulary size $V$, is one of the primary reasons for the huge memory footprint of embedding matrices. It is common to represent a token, $x$ by one-hot representation, $\mathbb{Y}(x) \in [0,1]^{V}$ and a distributed representation of the token is obtained by multiplying the one-hot representation with the embedding matrix, $W_{V} \in \mathbb{R}^{d \times V}$ as in 
$${U_{V}}(x) =  W_{V} * \mathbb{Y}(x)^\top \in \mathbb{R}^{d}$$


One way to remove the dependency on
the vocabulary size is to learn a smaller matrix, $W_{K} \in \mathbb{R}^{d \times K}$  ($K << V$), as shown in Figure \ref{fig:word_embedding_table_vs_lsh}. For instance, 300-dimensional Glove embeddings, $W_V$ \citep{pennington2014glove} with 400k vocabulary size occupies $>$ 1 GB while the $W_K$ occupies only $\approx$ 1.2 MB for $K=1000$ yielding a $1000 \times$ reduction in size.
Instead of learning a unique vector for each token in the vocabulary, we can think of the columns of this $W_{K}$ matrix as a set of basis vectors and each token can be represented as a linear combination of basis vectors in $W_{K}$ as in Figure \ref{fig:word_embedding_table_vector}. We select the basis vectors from $W_{K}$ for each token with a fixed $K$-bit binary vector instead of a $V$-bit one-hot vector. 

The LSH \textit{Projection function}, $\mathbb{P}$ (Figure \ref{fig:lsh_proj})\citep{projection_net2017,pmlr-v97-ravi19a} used in SGNN \citep{sgnn_emnlp18} and ProSeqo \citep{proseqo2019} does exactly this as it dynamically generates a fixed binary projection representation, $\mathbb{P}(x) \in [0,1]^{K}$ for any token, $x$ by extracting morphological input features like char (or token) n-gram \& skip-gram features, parts of speech tags etc.\ from $x$  and a modified Locality-Sensitive Hashing (LSH) based transformation, $\mathbb{L}$ as in $$x \xrightarrow{\mathbb{F}} [f_1, \cdots, f_n] \xrightarrow{\mathbb{L}} \mathbb{P}(x) \in [0,1]^{K}$$
where $\mathbb{F}$ extracts n-grams (or skip-grams), $[f_1, \cdots, f_n]$ from the input text. Here, $[f_1, \cdots, f_n]$ could refer to either character level or token level n-grams(or skip-grams) features. Given the LSH projection representation, $\mathbb{P}(x)$, the distributed representation of the token, $x$ is represented as in $${U_{W}}(x) = W_{K} * \mathbb{P}(x)^\top \in \mathbb{R}^{d}$$.
It is worth noting that projection operation, $\mathbb{P}$ can also be used map an entire sentence directly to the $[0,1]^K$ space.

As for the projection based classifiers, the ProSeqo model \citep{proseqo2019} runs a BiLSTM over word-level binary LSH projection representations to predict the correct classes, while the SGNN model \citep{sgnn_emnlp18} computes a binary LSH projection representation for the entire input text, followed by a 2-layer MLP and a softmax layer on top of it for class prediction.  SGNN was designed for short text, while ProSeqo is also suitable for long text classification tasks.

There have been a number of research efforts \citep{misspelling2, misspelling1, bertAdverserial2019} to improve the robustness of neural classifiers to misspelling attacks and other text transformations. 
Recently, \citet{bertAdverserial2019} observe that fine-tuned BERT and BiLSTM based models are very brittle (for e.g., accuracy drops from 90.3\% to 45.8\% in the SST \citep{socher2013recursive} classification task) to adversarial misspelling attacks. Contrary to intuition, they observe that  word-piece and character-level models are more susceptible to spelling attacks compared to the word-level models.

The LSH projection operation, $\mathbb{P}$, is a function of n-grams (and skip-grams) of the input text $x$ and usually the fraction of n-grams affected by spelling attacks tend to be minimal resulting in insignificant changes to the projection representation, $\mathbb{P}(x)$. 
Therefore, we hypothesize that the projection based models like ProSeqo, SGNN, etc. are \textit{inherently} robust to commonly occurring spelling attacks. 
In the following sections, we investigate the robustness of projection based classifiers by subjecting them to common misspellings, followed by an analysis of changes in the binary LSH projections of input text under such transformations.

\section{Effect of Misspellings on Text Classification} \label{sec:adversarial_attack}
We study the robustness of two types of projection based models -- ProSeqo and SGNN.
On the other hand, we fine-tune the pretrained BERT-base model (with word-piece tokenization) \citep{bert2018} and train two-layer BiLSTMs (with both word-only and word-piece tokenization) for comparable accuracies with respect to the projection based models for a fair comparison.
By word-only tokenization, we mean that models encode input words using a lookup table for each word. In our setup, we test the robustness of the neural classifiers by subjecting the corresponding test sets to common misspellings and omissions. We consider the following perturbation operations: randomly dropping, inserting, and swapping internal characters within words of the input sentences \citep{char_perturbation, bertAdverserial2019} \footnote{Further details on the perturbation operations and training details necessary for reproducibility are presented in the supplementary material}. We decide to perturb each word in a sentence with a fixed probability,  $P_{perturb}$. Following \citep{sgnn_emnlp18}, we fix the projection dimension to $K = 1120$. 

\subsection{Datasets}
For evaluation purposes, we use the following text classification datasets for dialog act classification MRDA \citep{MRDA} and SWDA \citep{Godfrey:1992:STS:1895550.1895693,Jurafsky}, for intent prediction ATIS \citep{atis} and long text classification Amazon Reviews \cite{zhang2015text} and Yahoo! Answers \cite{zhang2015text}.  Table \ref{table:datasets} shows the characteristics of each dataset. 

\begin{table}[!htbp]
\scalebox{0.61}{
\begin{tabular}{lcccc}
\hline
\textbf{Tasks} & \textbf{\# Classes} & \textbf{Avg-len} & \textbf{Train} & \textbf{Test}\\
\hline
ATIS (Dialog act) & $21$ & $11$ & $4.4k$ & $0.89k$ \\
MRDA (Dialog act) & $6$ & $8$ & 78k & 15k\\
SWDA (Intent Prediction) & 42 & 7 & 193k & 5k \\
YAHOO (Answers Categorization) & 10 & 108 & 1400k & 60k \\
AMAZON (Review Prediction) & 5 & 92 & 3000k & 650k \\  
\hline
\end{tabular}
}
\caption{Classification Dataset Characteristics}
\label{table:datasets}
\end{table}

\begin{table}[!h]
\scalebox{0.7}{
\begin{tabular}{l|cccc}
\hline
\multicolumn{5}{c}{\textbf{Accuracy drop} (\%) $\pm$ \textbf{std-deviation} (\textbf{over 5 runs})}\\
\hline
\textbf{Datasets} $\rightarrow$ & \textbf{MRDA} & \textbf{ATIS} & \textbf{YAHOO} & \textbf{AMAZON}\\
\textbf{Models} $\downarrow$ & & & & \\
\hline
BERT-base & $8.25_{\pm3.4}$ & $15.57_{\pm1.2}$ & $9.06_{\pm5.4}$ & $12.88_{\pm3.8}$\\
BiLSTM-wp & $8.91_{\pm3.9}$ & $20.11_{\pm3.1}$ & $11.32_{\pm4.5}$ & $9.46_{\pm1.3}$\\
BiLSTM-w & $15.14_{\pm4.3}$ & $16.23_{\pm2.6}$ & $9.32_{\pm2.3}$ & $8.88_{\pm2.4}$\\
SGNN & $\mathbf{1.91_{\pm0.5}}$ & $\mathbf{2.80_{\pm0.3}}$ & - & -\\
ProSeqo & $2.11_{\pm0.4}$ & $2.84_{\pm0.6}$ & $\mathbf{3.11_{\pm0.3}}$ & $\mathbf{3.91_{\pm0.5}}$\\
\hline
\end{tabular}
}
\caption{Each entry in the table denotes the average drop in accuracy(\%) when classifiers are subject to test inputs with misspellings, $P_{perturb} = 0.2$. ATIS and MRDA are smaller datasets, while Yahoo! Answers and Amazon reviews are larger. BERT-base refers to the BERT-base \citep{bert2018} word-piece model fine-tuned to individual tasks and it converged to state-of-the-art for all tasks.}
\label{table:bert_lstm_proseqo_cmp}
\end{table}

\begin{table*}[!ht]
\centering
\scalebox{0.68}{
\begin{tabular}{c|ccccccccc}
\hline
\multicolumn{9}{c}{\textbf{Accuracy}(\%) \textbf{(Averaged over 5 runs)}}\\
\hline
\textbf{Datasets} & \multicolumn{3}{c|}{\textbf{MRDA}} & \multicolumn{3}{c|}{\textbf{ATIS}} & \multicolumn{3}{c}{\textbf{SWDA}}\\
\hline
\textbf{Perturb}(\%) & \textbf{BiLSTM-wp} & \textbf{BiLSTM-w} & \textbf{SGNN} & \textbf{BiLSTM-wp} & \textbf{BiLSTM-w} & \textbf{SGNN} & \textbf{BiLSTM-wp} & \textbf{BiLSTM-w} & \textbf{SGNN}\\
\hline
$0$ & $79.23$ & $78.14$ & $87.22$ & $91.73$ & $92.04$ & $93.51$ & $72.92$ & $72.94$ & $76.21$ \\
\hline
\multicolumn{9}{c}{Perturbation operation: \textit{drop}}\\
\hline
$20$ & $69.46_{\pm1.1}$ & $74.96_{\pm0.7}$ & $85.43_{\pm0.3}$ & $81.95_{\pm2.2}$ & $80.15_{\pm2.0}$ & $91.05_{\pm0.2}$ & $65.05_{\pm4.2}$ & $64.95_{\pm4.1}$ & $70.76_{\pm1.1}$\\

$40$ & $72.34_{\pm2.9}$ & $55.17_{\pm3.1}$ & $84.62_{\pm0.22}$ & $71.69_{\pm2.9}$ & $65.88_{\pm2.1}$ & $91.86_{\pm0.3}$ & $61.24_{\pm4.1}$ & $64.95_{\pm5.2}$ & $67.79_{\pm1.2}$\\

$60$ & $69.81_{\pm4.7}$ & $42.25_{\pm4.2}$ & $83.27_{\pm0.25}$ & $59.16_{\pm4.8}$ & $56.25_{\pm3.9}$ & $90.12_{\pm0.4}$ & $57.48_{\pm5.7}$ & $58.77_{\pm5.8}$ & $63.21_{\pm1.3}$\\
\hline
\multicolumn{9}{c}{Perturbation operation: \textit{swap}}\\
\hline
$20$ & $78.25_{\pm1.1}$ & $71.34_{\pm2.2}$ & $86.74_{\pm0.1}$ & $86.12_{\pm1.7}$ & $85.05_{\pm1.8}$ & $92.05_{\pm0.3}$ & $66.27_{\pm3.3}$ & $64.52_{\pm2.1}$ & $70.84_{\pm0.3}$\\

$40$ & $75.91_{\pm3.9}$ & $69.22_{\pm2.1}$ & $86.39_{\pm0.2}$ & $82.06_{\pm2.7}$ & $78.04_{\pm2.9}$ & $91.15_{\pm0.2}$ & $62.67_{\pm5.8}$ & $54.93_{\pm2.4}$ & $67.22_{\pm0.3}$\\

$60$ & $69.22_{\pm3.8}$ & $66.91_{\pm4.0}$ & $85.99_{\pm0.2}$ & $72.34_{\pm3.3}$ & $68.54_{\pm4.1}$ & $91.27_{\pm0.3}$ & $59.20_{\pm4.8}$ & $47.93_{\pm4.3}$ & $64.48_{\pm0.3}$\\
\hline
\multicolumn{9}{c}{Perturbation operation: \textit{all}}\\
\hline
$20$ & $72.96_{\pm1.3}$ & $73.39_{\pm2.3}$ & $86.71_{\pm0.4}$ & $80.40_{\pm1.7}$ & $83.55_{\pm2.1}$ & $92.83_{\pm0.2}$ & $60.49_{\pm4.3}$ & $64.28_{\pm3.3}$ & $68.96_{\pm0.2}$\\

$40$ & $70.32_{\pm3.4}$ & $63.04_{\pm4.3}$ & $85.31_{\pm0.5}$ & $71.62_{\pm3.1}$ & $75.81_{\pm2.6}$ & $90.71_{\pm0.3}$ & $54.96_{\pm4.1}$ & $59.46_{\pm4.6}$ & $65.44_{\pm0.4}$\\

$60$ & $67.64_{\pm5.7}$ & $55.50_{\pm5.3}$ & $84.21_{\pm0.5}$ & $61.10_{\pm6.1}$ & $66.58_{\pm5.6}$ & $88.35_{\pm0.3}$ & $49.62_{\pm6.7}$ & $51.85_{\pm6.3}$ & $64.97_{\pm0.5}$\\
\hline
\end{tabular}

}
\caption{Comparison of projection based models vs BiLSTMs subject to various types and amounts of perturbations. BiLSTM-wp and BiLSTM-w refer to models with word-piece and word-only tokenization respectively.}
\label{table:Adversarial_results}
\end{table*}

\subsection{Experiments and Results}
Table \ref{table:bert_lstm_proseqo_cmp} reports the average classifier accuracy drops when all the models are subject to all types of perturbations (\textit{swap}, \textit{drop}, \& \textit{add}) on multiple classification tasks (two short text and two long text). We see that the accuracy drop for the projection based models is significantly lower across all datasets. It is also worth noting that the standard deviations across the $5$ runs are also minimal for the projection based models further showcasing the stability of projection representations.

In another experiment shown in Table \ref{table:Adversarial_results}, we subject different models to varying types and amounts of perturbations. Similarly, we see that the accuracy drop for the projection based models is the smallest across all datasets and amounts of perturbation. Compared to the word-only models, we observe that the word-piece models are also comparably susceptible to character perturbations which agrees with the findings in \citep{bertAdverserial2019}.

\section{Perturbation Analysis}
\label{sec:perturbation_study}
Apart from the classification experiments, we also directly analyze the changes in the binary LSH projection representations by subjecting input text to different types and amount of perturbations.
To that end, we take a large corpus \textit{\textbf{enwik9}}\footnote{\textit{enwik9} is a byte-level dataset consisting of the first $10^9$ bytes of the English Wikipedia XML dump,  http://mattmahoney.net/dc/textdata.html.} (vocabulary size of $500 k$ and $129 M$ words) to analyze the average Hamming distance between LSH projections of the words in the corpus. Next, we compute the average changes in the projection representations by subjecting them to the character perturbations from Section \ref{sec:adversarial_attack}. Table \ref{table:Pertbation_results_word} shows the results. We make the following observations from our experiments:
\begin{enumerate}
    \item Average Hamming distance between LSH projections of words is  $\approx K/2$, where $K$ is the projection dimension which implies that the words are more or less uniformly spread out from each other indicating that there are no bias issues in the $[0,1]^{K}$ representation space. 
    
    \item Assuming $P_{perturb} = 0.2$, we observe that LSH projection changes only by $\approx$ $11$\% w.r.t the average Hamming distance between the words in the corpus when subject to misspellings. For instance, if the average Hamming distance between LSH projections of words is $100$ bits, misspellings change the projections by only $11$ bits on average. Intuitively, this suggests that neural layers on top of the LSH projection tend to rarely confuse a misspelled word for another valid word.     
\end{enumerate}
Also from Table \ref{table:Pertbation_results_word}, we found that the changes in the LSH-projection, $\Delta_{\mathbb{P}(x)}$ due to perturbations is directly proportional to LSH projection dimension, $K$ and perturbation probability, $P_{perturb}$ as in, $\Delta_{\mathbb{P}(x)} \propto K\, \cdot P_{perturb}$.

\begin{table}[t!]
\centering
\scalebox{0.60}{
\begin{tabular}{c|cc}
\hline
\textbf{LSH Proj.Dim} ($\mathbf{K}$) & \multicolumn{2}{c}{\textbf{Character Perturbations}}\\
  & 5\% & 10\% \\
\hline
840 & 10.08 & 24.33\\
980 & 15.48 & 31.24\\
1120 & 18.83 & 33.65\\
1260 & 19.71 & 39.01\\
\hline
\end{tabular}
}
\caption{Avg. changes in word projections (bits) for different Char Perturbation \% in \textit{enwik9} corpus.}
\label{table:Pertbation_results_word}
\end{table}

\section{Conclusion}
In this work, we perform a detailed study analyzing the robustness of recent LSH-based projection neural networks for memory-efficient text representations. Based on multiple text classification tasks and perturbation studies, we find projection-based neural models to be robust to text transformations compared to BERT or BiLSTMs with embedding lookup tables for words and word-pieces.

\bibliography{anthology,eacl2021}
\bibliographystyle{acl_natbib}

\end{document}


\maketitle

\section{Supplemental Material}
\label{sec:supplemental}

\subsection{Perturbation Operations}
We mainly experiment with three types of character level perturbation operations inspired from the techniques used in \citep{char_perturbation}.
\begin{itemize}
\item \textit{add(word, n)} : We randomly choose \textit{n} characters from the character vocabulary and insert them at random locations into the input \textit{word}. We however retain the first and last characters of the word as is. Example transformation: $sample \rightarrow samnple$.

\item \textit{swap(word, n)}: We randomly swap the location of two characters in the word \textit{n} times. As with the \textit{insert} operation, we retain the first and last characters of the word as is and only apply the \textit{swap} operation to the remaining characters. Example transformation: $sample \rightarrow sapmle$.

\item \textit{drop(word, n)}: We randomly drop \textit{n} character in the word by \textit{n} times. Example transformation: $sample \rightarrow smple$.

\end{itemize}

We decide to perturb each word in a sentence with a fixed probability, $P_{perturb}$. In all our experiments, we set $n=1$ and projection dimension to K = 1120, following \citep{sgnn_emnlp18}. For the perturbation operations, we explore experiments with $P_{perturb} = {0.2, 0.4, 0.6}$. 

\subsection{Training Details}
We ran our experiments in individual Nvidia Tesla-p100 GPUs and optimized using the ADAM optimizer \citep{adam} with the default hyper-parameters used in \citep{salasforce_lm1,salesforce_lm2}. 
All the classifiers mentioned in the paper are trained with batch size $32$, a learning rate $0.0001$ and early stopping with patience parameter of $10$. 

For the BERT \citep{bert2018} classifiers in Table 2, we fine-tune the publicly available, BERT-BASE (L=$12$, H=$768$, A=$12$, Total Parameters=$110M$) \footnote{https://github.com/google-research/bert} model for all the four datasets.

For the two-layer BiLSTMs-based classifiers in Table 3, we set the embedding-size=$512$, hidden-size=$1024$ and sentence cutoff length=$80$ for all the datasets. For the word-piece variation, we employ the byte level bpe tokenizer from hugging face, with a vocabulary size of $8k$\footnote{https://github.com/huggingface/tokenizers}.

\bibliography{anthology,eacl2021}
\bibliographystyle{acl_natbib}